\documentclass{article}
\usepackage{microtype}
\usepackage{booktabs} 

\usepackage{hyperref}

\usepackage{graphicx}
\usepackage{comment}
\usepackage{amsmath,amssymb} 
\usepackage{color}
\usepackage[caption=false]{subfig}
\usepackage{booktabs}
\usepackage{multirow}
\usepackage{threeparttable}
\usepackage{bm}
\usepackage[normalem]{ulem}

\usepackage[accepted]{icml2020}

\icmltitlerunning{A Critical Evaluation of Open-World Machine Learning}

\begin{document}

\twocolumn[
\icmltitle{A Critical Evaluation of Open-World Machine Learning}



\icmlsetsymbol{equal}{*}

\begin{icmlauthorlist}
\icmlauthor{Liwei Song}{equal,princeton}
\icmlauthor{Vikash Sehwag}{equal,princeton}
\icmlauthor{Arjun Nitin Bhagoji}{equal,princeton}
\icmlauthor{Prateek Mittal}{princeton}
\end{icmlauthorlist}

\icmlaffiliation{princeton}{Princeton University}
\icmlcorrespondingauthor{Liwei Song}{liweis@princeton.edu}

\icmlkeywords{Machine Learning, ICML}

\vskip 0.3in
]



\printAffiliationsAndNotice{\icmlEqualContribution} 

\newcommand{\dummytext}[1]{{\color{white} #1}}
\newcommand{\bluetext}[1]{{\color{blue} #1}}

\newcommand{\vikash}[1]{{\color{cyan} Vikash: #1}}
\newcommand{\arjun}[1]{{\color{blue} Arjun: #1}}
\newcommand{\liwei}[1]{{\color{green} Liwei: #1}}
\newcommand{\todo}[1]{{\color{red} To-Do: #1}}

\newcommand{\bcomment}[1]{}

\newcommand{\oodAdvExamples}[0]{OOD adversarial examples\xspace}    
\newcommand{\oodattack}[0]{OOD attack} 
\newcommand{\oodattacks}[0]{OOD attacks }
\newcommand{\oodlong}[0]{out-of-distribution }
\newcommand{\ood}[0]{OOD }
\newcommand{\intphoto}[0]{Internet Photos }

\newcommand{\pgdx}[0]{\textsf{PGD-xent}}
\newcommand{\pgdcw}[0]{\textsf{PGD-CW}}

\newcommand{\V}[1]{\mathbf{#1}}
\newcommand{\bfC}{\V{C}}
\newcommand{\bfu}{\V{u}}
\newcommand{\bfU}{\V{U}}
\newcommand{\bfw}{\V{w}}
\newcommand{\bfv}{\V{v}}
\newcommand{\bfX}{\V{X}}
\newcommand{\bfx}{\V{x}}
\newcommand{\xood}{\bfx_{\text{OOD}}}
\newcommand{\bfz}{\V{z}}
\newcommand{\bfy}{\V{y}}
\newcommand{\bfZ}{\bm{\phi}}
\newcommand{\xad}{\tilde{\bfx}}
\newcommand{\xadood}{\tilde{\bfx}_{\text{OOD}}}

\newcommand{\mcX}{\mathcal{X}}
\newcommand{\mcY}{\mathcal{Y}}
\newcommand{\R}{\mathbb{R}}
\begin{abstract}
Open-world machine learning (ML) combines closed-world models trained on in-distribution data with out-of-distribution (OOD) detectors, which aim to detect and reject OOD inputs.
Previous works on open-world ML systems usually fail to test their reliability under diverse, and possibly adversarial conditions.
Therefore, in this paper, we
seek to understand \emph{how resilient are state-of-the-art open-world ML systems to changes in system components?}
With our evaluation across 6 OOD detectors, we find that the choice of in-distribution data, model architecture and OOD data have a strong impact on OOD detection performance, inducing false positive rates in excess of 70\%.
We further show that OOD inputs with 22 unintentional corruptions or adversarial perturbations render open-world ML systems unusable with false positive rates of up to 100\%.
To increase the resilience of open-world ML,
we combine robust classifiers with OOD detection techniques and uncover a new trade-off between OOD detection and robustness.
\end{abstract}

\section{Introduction} \label{sec: introduction}

Open-world machine learning (ML) \cite{geng2018opensetreview} has been proposed to deal with arbitrary inputs to real-world ML systems, which may not be drawn from the distribution used during training, referred to as in-distribution data ($D_{\text{in}}$).
In this paradigm, \emph{out-of-distribution (OOD) detectors}~\cite{hendrycks2017iclr} are added to existing supervised learning systems to detect OOD examples, with classification outputs only provided for those inputs determined to come from $D_{\text{in}}$, and others marked as anomalous. 

Several OOD detection mechanisms~\cite{liang2017ODIN,dhamija2018agnostophobia,lee2018unifieddetector} have been recently proposed which exhibit good performance in certain settings. However, they fail to test the resilience of open-world ML systems. For example, previous work typically uses only a few OOD datasets to evaluate OOD detectors along with a fixed $D_{\text{in}}$ and model architecture.
Further, they only use clean OOD datasets for evaluation without considering cases when OOD data is corrupted or adversarially perturbed.

We seek to overcome this limitation and ask:  \emph{how resilient and effective are current OOD detectors?} We categorize and evaluate 6 key OOD detectors from top machine learning conferences, in a unified evaluation setup which varies $D_{\text{in}}$, the OOD dataset, and the classifier architecture. We find that the use of deep feature representations aids OOD detection, but when faced with specially curated, diverse OOD data, most OOD detectors we test have very high false positive rates.
We also test OOD detectors against images with 22 types of corruptions, which could either improve or decrease the detection performance significantly.
Built on our recent analysis \cite{sehwag2019analyzing} of attacking the classifier with OOD data, we finally propose a method to generate \ood adversarial examples against end-to-end open-world ML systems and demonstrate that it has a close to 100\% success rate in most cases.

Given this worrying impact of OOD adversarial examples, we integrate approaches for robust and open-world ML to investigate \emph{how plausible the development of robust open-world ML systems is}. We adapt adversarial training \cite{szegedy2014intriguing,madry_towards_2017}, a promising approach for robustness against adversarial examples generated from $D_{\text{in}}$, to work with various OOD detection mechanisms. We show that while this approach has some promise, it introduces a new trade-off between OOD detection performance and robustness. We find it arises from a reduction in the discriminatory power of robust features for OOD detection.

\section{Open World Machine Learning: Background and Related Work} \label{sec: background}

\label{subsec: open_world_ml}
In open-world ML \cite{geng2018opensetreview}, inputs to the classifier $f(\cdot)$ at test time is not limited to be from the same distribution $D_{\text{in}}$ used for training. They can be \emph{out-of-distribution (OOD)}. 
To handle OOD data, an \ood detector $h(\cdot)$ is usually used to ensure that \ood data is marked as such and not passed on the classifier.
Here, we provide brief descriptions of 6 state-of-the-art OOD detectors we analyze and critique in this paper. These represent a diversity of techniques adopted to deal with OOD data, 
particularly regarding their dependence on the classifier $f(\cdot)$ 
and need for representative OOD data, either for training or calibration.

\noindent \textbf{ODIN} \cite{liang2017ODIN} (\emph{Depends on $f(\cdot)$}-Yes, \emph{Use of OOD data}-Calibration): This detector thresholds the maximum class-wise output confidence of a classifier and if it is below a determined threshold, then the input is classified as being OOD. The threshold may be determined with or without the use of some \ood data. In addition, each input is slightly perturbed to increase its predicted confidence and the classifier's output vector is temperature scaled. \\
\noindent \textbf{Network Agnostophobia} \cite{dhamija2018agnostophobia} (\emph{Depends on $f(\cdot)$}-Yes, \emph{Use of OOD data}-Training): This detector uses the output confidence of a neural network trained with an additional loss term which maximizes the entropy of the softmax output on OOD data. To perform this training, generic \ood data is used during training (e.g., using CIFAR-100 when CIFAR-10 is in-distribution). \\
\textbf{Mahalanobis Detector} \cite{lee2018unifieddetector} (\emph{Depends on $f(\cdot)$}-Yes, \emph{Use of OOD data}-Calibration): Class-conditional Gaussian distributions are first derived from the feature representations of a classifier on training data. Then, for each input, the Mahalanobis distance between the its features and the nearest class-conditional distribution is computed. An input with a Mahalanobis distance larger than the threshold is inferred as OOD. Some \ood data is used for threshold calibration and weights calibration for distance calculation.\\
\textbf{Autoencoder Detector} \cite{chalapathy2017robustAE} (\emph{Depends on $f(\cdot)$}-No, \emph{Use of OOD data}-None): A separate autoencoder is trained using the training data to minimize the reconstruction error. OOD data is likely to have different latent representations, which lead to larger reconstruction errors as a sign of OOD. In this paper, we use the denoising autoencoder structure used by \citet{meng2017magnet}.\\
\textbf{Deep-SVDD} \cite{ruff2018deepsvm} (\emph{Depends on $f(\cdot)$}-No, \emph{Use of OOD data}-None): In this approach, a support vector data description (SVDD) is trained on features extracted from a base network pre-trained using an autoencoder. The SVDD then bounds the in-distribution data with a hypersphere and classifies inputs outside it as OOD. \\
\textbf{Outlier Exposure} \cite{hendrycks2018outlierexpose} (\emph{Depends on $f(\cdot)$}-Yes, \emph{Use of OOD data}-Training): Similar to Network Agnostophobia, this detector also aims to improve the \ood detection performance of the networks by explicitly training on out-of-distribution inputs by adding an OOD loss term (usually the entropy of the output probability vector).

\subsection{Connecting robust and open-world ML} \label{subsec: robust_ml}
Our prior work \cite{sehwag2019analyzing} was the first to connect robust and open-world ML through the introduction of OOD adversarial examples, which were limited to inducing target misclassification in the classifier. Compared to \citet{sehwag2019analyzing}, here we (i) consider the effect of network architecture and $D_{\text{in}}$ on \ood detection, finding a large variation in the 6 detectors we test; (ii) consider far more diverse OOD data, including corruptions and end-to-end adversarial examples; (iii) combine \ood detectors with robust training to reveal a tradeoff between OOD detection and robustness.

\section{Failure Modes of Open-World ML}\label{sec: evaluation}

Previous work proposing OOD detection techniques has been limited in its evaluation. We analyze the resilience of these OOD detectors to changes in the OOD data, model architecture, and $D_{\text{in}}$, finding a large variation in performance as these aspects are changed. We then analyze the robustness of open-world ML systems under  22 types of image corruptions \cite{hendrycks2019natural} to \ood data and end-to-end \ood adversarial examples.

\noindent \textbf{In-distribution data:} We primarily use CIFAR-10 as in-distribution data for training ($D_{\text{in}}$), given its prevalence in previous work on \ood detection. We also report results with Tiny-Imagenet as $D_{\text{in}}$ to evaluate the impact of $D_{\text{in}}$.

\noindent \textbf{Models:} We use a Wide ResNet \cite{zagoruyko2016wide} with width and depth equal to 10 and 28 (WRN-28-10) as the primary classifier. We also report results with DenseNet-100, to evaluate the impact of model architecture.

\noindent \textbf{Metrics:} We report the false positive rate (FPR) to measure the detection performance, which is the percentage of OOD examples that are not detected. Following previous work, we report our results with the OOD detector calibrated to achieve a $95\%$ true positive rate (TPR), i.e., $95\%$ of in-distribution data passes through the detector successfully.

\subsection{Effect of OOD data}\label{subsec: data_effect}

We consider 6 OOD datasets of two types.
(i) \emph{Semantically meaningful OOD data}: besides the public MNIST, VOC12 and Imagenet datasets, we also use the \intphoto dataset \cite{sehwag2019analyzing} which contains 1000 distinct natural images from the internet using the Picsum service \cite{picsum} and was not used in previous evaluations of OOD detectors. We ensure the label set of \ood examples is distinct from that of $D_{\text{in}}$.
(ii) \emph{Noise OOD data}: we construct Gaussian and Uniform Noise datasets, consisting of random points in the input space.
We scale all OOD examples to the dimension of in-distribution data. 
All results are obtained with 1,000 images from each dataset.

\begin{table}[t]
	\caption{FPR (lower is better) of OOD detectors for 6 different OOD datasets at a 95\% TPR. \textbf{Model:} WRN-28-10, \textbf{In-distribution:} CIFAR-10.}
	 	 \label{tab: benign_data_var_resnet}
	    \resizebox{\linewidth}{!}{
		\begin{tabular}{c|c|c|c|c|c|c}
		    \toprule
            \multirow{2}{*}{OOD Detector}  & \multirow{2}{*}{MNIST} & \multirow{2}{*}{VOC12} & \multirow{2}{*}{Imagenet} & Internet  & Gaussian  & Uniform  \\ 
              &  &  &  &  Photos &  Noise &  Noise \\ \midrule
            ODIN & 28.2  & 17.8 & 24.7 & 30.3 & 13.3 & 12.1 \\ \hline
            Network Agnostophobia & 15.6  & 25.3 & 22.2 &  \textbf{14.0} & \textbf{0.0} & \textbf{0.0}\\ \hline
            Mahalanobis Detector & 8.9  & \textbf{10.8}  & \textbf{7.3} & 76.5 & \textbf{0.0} & \textbf{0.0} \\ \hline
            Autoencoder Detector & \textbf{0.0}  & 56.1 & 53.5 & 93.6 & \textbf{0.0} & \textbf{0.0} \\ \hline
            Deep-SVDD & 59.8 & 51.6 & 58.6 & 97.9 & 18.4 & \textbf{0.0} \\ \hline
            Outlier Exposure & 18.5 & 23.4 & 31.1 & 24.6 & \textbf{0.0} & \textbf{0.0} \\
            \bottomrule
        \end{tabular}
	}
	\vspace{-18pt}
\end{table}

\noindent \textbf{Use of classifier's feature representations aids OOD detection:} In Table \ref{tab: benign_data_var_resnet}, we see that compared to Classifier-independent detectors (Autoencoder and Deep-SVDD), OOD detectors relying on the classifier's deep feature representations achieve much lower FPR values in general. 

\noindent \textbf{Detectors fail on diverse OOD data:} The \intphoto dataset demonstrates the pitfalls of relying only on limited datasets to evaluate the performance of OOD detectors. 
The Mahalanobis detector outperforms the other detectors on most OOD data except for the \intphoto data, where it has a very high FPR, making it unusable at 95\% TPR.

\subsection{Effect of model architecture}

\begin{table}[t]
	\caption{FPR of OOD detectors using the DenseNet architecture. \textbf{Model:} DenseNet-100, \textbf{In-distribution:} CIFAR-10. 
	Autoencoder and Deep-SVDD are independent of the classifier used, so have the same performance as in Table \ref{tab: benign_data_var_resnet}.}
	 	 \label{tab: benign_data_densenet}
	    \resizebox{\linewidth}{!}{
		\begin{tabular}{c|c|c|c|c|c|c}
		    \toprule
		    \multirow{2}{*}{OOD Detector}  & \multirow{2}{*}{MNIST} & \multirow{2}{*}{VOC12} & \multirow{2}{*}{Imagenet} & Internet  & Gaussian  & Uniform  \\ 
              &  &  &  &  Photos &  Noise &  Noise \\ \midrule
            ODIN & 6.0  & 24.9 & 29.2 & 41.5 & 9.9 & 53.3 \\ \hline
            Network Agnostophobia & 39.3 & 28.6 & 25.8 & \textbf{25.3} & \textbf{0.0} & \textbf{0.0} \\ \hline
            Mahalanobis Detector & 23.0  & \textbf{10.9} & \textbf{8.3} & 73.9 & \textbf{0.0} & \textbf{0.0} \\ \hline
            Outlier Exposure & 21.7 & 32.1 & 36.6 & 32.3 & 99.5 & 85.7 \\
            \bottomrule
        \end{tabular}
	}
	\vspace{-18pt}
\end{table}

\noindent \textbf{Change in model architecture can cause drastic performance dips:} As shown in Table \ref{tab: benign_data_densenet}, by using DenseNet-100, ODIN achieves lower FPR on MNIST and Gaussian Noise but higher FPR on other datasets. For Network Agnostophobia, Mahalonobis detector, and Outlier Exposure, the OOD detection performance gets worse. Outlier Exposure even has the FPR higher than $85\%$ on Noise OOD data.

\subsection{Effect of $D_{\text{in}}$}
We use Tiny-Imagenet as $D_{\text{in}}$ to study the impact of $D_{\text{in}}$ on OOD detectors.
Detailed results are in the Appendix \ref{appendix:tiny-imagenet}.

\noindent \textbf{OOD detectors have poor performance with Tiny-Imagenet as $D_{\text{in}}$:} We observe a drastic reduction in the detection performance when we switch $D_{\text{in}}$ to Tiny-Imagenet. Similar behavior was also observed in \citet{hendrycks2018outlierexpose}. For example, with CIFAR-10 as $D_{\text{in}}$, ODIN achieves less than 15\% FPR for Gaussian and Uniform noise with WRN-28-10 model. However, when using Tiny-Imagenet as $D_{\text{in}}$, the FPR increases to 100\%. We observe similar phenomena for other OOD datasets.

\subsection{Sensitivity to corrupted data}
We measure the performance of ODIN and Mahalanobis detector \footnote{These are chosen for having the overall best performance in the previous evaluation.} under 22 distinct types of corruptions (details are in Appendix \ref{appendix:corruptions}), which we categorize as Geometric, Blurring, Noise, Photometric and Weather based on their salient characteristics. 
Intuitively, we expect that any corruption to an image would make it easier to detect as OOD, since the corruption is unlikely to be encountered in sanitized $D_{\text{in}}$.

\begin{figure}[t]
	\centering
	\subfloat[ODIN detector]{\resizebox{0.3\textwidth}{!}{\includegraphics[width=0.5\linewidth]{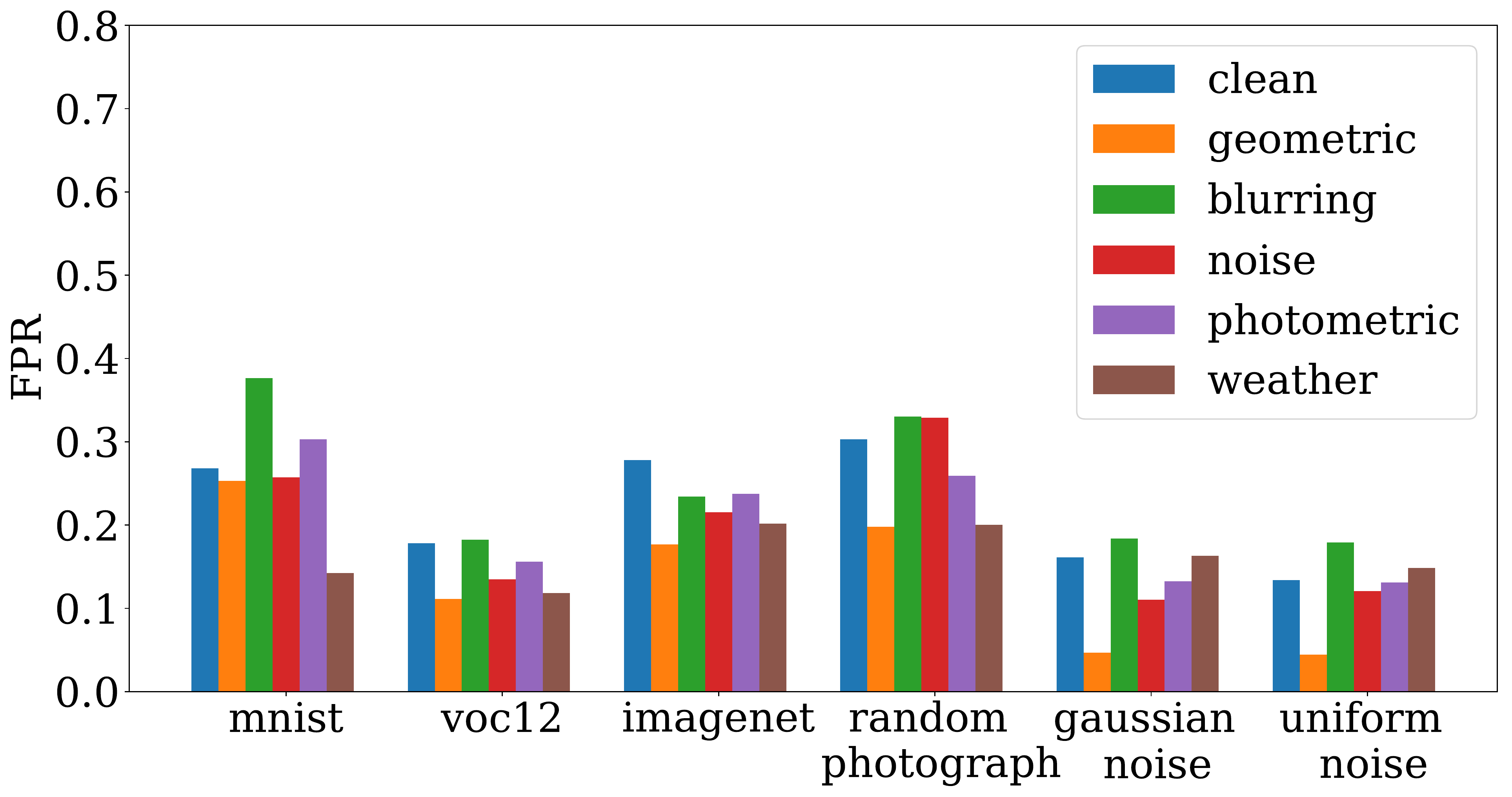}}\label{subfig: ODIN_corruption}}
	\hspace{0mm}
	\subfloat[Mahalanobis detector]{\resizebox{0.3\textwidth}{!}{\includegraphics[width=0.5\linewidth]{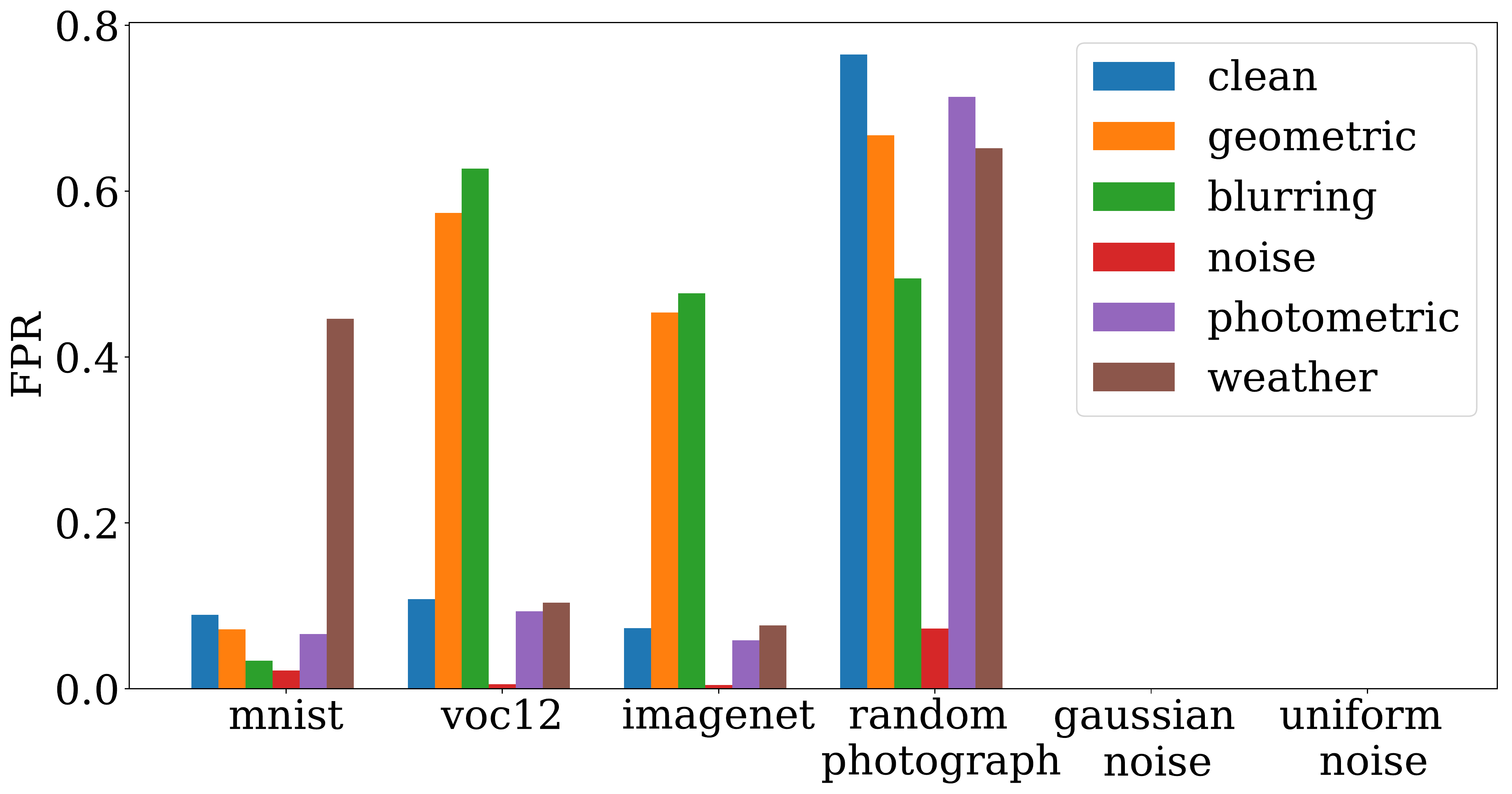}}\label{subfig: Mahalanobis_corruption}}
	\caption{FPR for OOD detectors on corrupted OOD data}
	\label{fig: cifar_corruption}
	\vspace{-15pt}
\end{figure}

\noindent \textbf{Corruptions can increase or decrease FPR for OOD detectors:} Contrary to what would be intuitively expected, OOD detectors have different detection behavior depending on the type of image corruptions and even the OOD dataset. Both detectors have considerable variation in their performance with the addition of corruptions (Fig. \ref{fig: cifar_corruption}). For the Mahalanobis detector (Fig. \ref{subfig: Mahalanobis_corruption}), Geometric and Blurring corruptions increase the FPR by around $5\times$ for VOC12 and Imagenet while leading to a modest decrease in FPR for MNIST and \intphoto datasets.

\subsection{Inducing targeted attacks in open-world ML}
We now consider the addition of adversarial perturbations to OOD data to induce targeted misclassification to entire open-world ML systems and assume that the attacker has knowledge of the entire system. 
The OOD detectors can be represented as first computing a distance metric for each input and then flagging the input as OOD if its distance is larger than a preset threshold.
The distance metric can be prediction uncertainty, or Euclidean distance of the input's representation, or reconstruction error.
Let $D_{h}(\cdot)$ represent the distance computation function for the detector $h$.
Now, to achieve targeted misclassification $T$ of an OOD input $\xadood$, the attacker needs to solve the following problem.
\begin{align}\label{eq: combined}
\min_{\bfx \in \mathcal{H}} \lambda \cdot D_{h}(\bfx) + \ell_{\text{xent}}(f(\bfx), T)
\end{align}
For our experiments, we use an $L_{\infty}$ constraint of $8$.
We run Projected Gradient Descent attacks \cite{madry_towards_2017} for $100$-$1000$ iterations and vary $\lambda$ to report best attack results. We also report the \emph{open-world target success rate (OW-TSR)}, which is the fraction of \ood adversarial examples that bypass the detector and achieve targeted misclassification. 

\begin{table}[t]
	\caption{FPR/OW-TSR (lower is better for both) with \emph{OOD adversarial examples} generated with an $L_{\infty}$ constraint of 8. \textbf{Model:} WRN-28-10, \textbf{In-distribution:} CIFAR-10.}
	\label{tab: adv_data_var_resnet}
	\resizebox{\linewidth}{!}{
		\begin{tabular}{c|c|c|c|c|c|c}
			\toprule
			\multirow{2}{*}{OOD Detector}  & \multirow{2}{*}{MNIST} & \multirow{2}{*}{VOC12} & \multirow{2}{*}{Imagenet} & Internet  & Gaussian  & Uniform  \\ 
			&  &  &  &  Photos &  Noise &  Noise \\ \midrule
			ODIN & 100.0/100.0   & 100.0/100.0 & 100.0/100.0 & 100.0/100.0 & 99.6/99.6 & 73.1/58.4\\ \hline
			Network Agnostophobia &  99.5/99.5 & 96.3/96.3 & 97.1/97.1 & 99.4/99.4 & 45.8/45.8 & 35.6/35.6\\ \hline
			Mahalanobis Detector & 94.1/93.6  & 95.4/93.6 & 88.7/87.1 & 97.3/96.8 & 0.0/0.0 & 0.0/0.0 \\ \hline
			Autoencoder Detector & 0.0/0.0 &  88.2/88.2  & 82.5/82.5 & 99.2/99.2  & 0.0/0.0  & 0.0/0.0 \\ \hline
			Deep-SVDD & 100.0/100.0  & 99.9/99.9 & 99.5/99.5 & 100.0/100.0 & 99.0/99.0 & 59.6/59.6 \\ \hline
			Outlier Exposure & 99.2/99.2 & 99.5/99.5 & 100.0/100.0 & 99.8/99.8 & 71.2/71.2 & 21.8/21.8 \\
			\bottomrule
		\end{tabular}
	}
	\vspace{-15pt}
\end{table}

\noindent \textbf{Results:} We present all attack results in Table \ref{tab: adv_data_var_resnet}.
We can see that \ood adversarial examples achieve both high FPR and OW-TSR values (around $100\%$) in most cases. This indicates the severe vulnerability of open-world ML systems. 

\section{Towards Robust Open-World ML} \label{sec: robust}

Adversarial training \cite{madry_towards_2017} has been shown to be successful in mitigating in-distribution adversarial examples.
Here, we study the impact of adversarial training on open-world ML systems by considering a PGD attacker with an $L_{\infty}$ constraint of $8$ and the WRN-28-10 model.

\begin{table}[t]
	\caption{FPR/OW-TSR (lower is better for both) with \emph{adversarial data} of OOD detectors with adversarial training for CIFAR-10 data for 6 different OOD datasets. \textbf{Model:} ResNet-28-10, \textbf{In-distribution:} CIFAR-10.}
	\label{tab: adv_data_var_resnet_adv_train}
	\resizebox{\linewidth}{!}{
		\begin{tabular}{c|c|c|c|c|c|c}
			\toprule
			\multirow{2}{*}{OOD Detector}  & \multirow{2}{*}{MNIST} & \multirow{2}{*}{VOC12} & \multirow{2}{*}{Imagenet} & Internet  & Gaussian  & Uniform  \\ 
              &  &  &  &  Photos &  Noise &  Noise \\ \midrule
			ODIN &  65.6/3.6 & 46.8/21.7 & 49.2/20.0 & 55.4/29.3 & 32.0/7.2 & 49.6/9.4\\ \hline
			Network Agnostophobia & 50.4/12.8  & 63.4/46.3 & 57.3/39.7 & 58.2/47.2 & 10.3/9.5 & 0.1/0.1\\ \hline
			Mahalanobis Detector &  10.6/4.2 & 86.8/32.0 & 47.9/41.0 & 90.4/38.4 & 0.0/0.0 & 0.0/0.0 \\ \hline
			Autoencoder Detector &  0.0/0.0  & 55.6/27.1 & 53.5/25.6 & 93.6/45.6 & 0.0/0.0 & 0.0/0.0 \\ \hline
			Deep-SVDD & 100.0/15.3  & 97.6/48.6 & 96.7/41.5 & 100.0/52.7 & 100.0/34.9 & 53.4/26.9\\ \hline
			Outlier Exposure & 64.6/2.3 & 62.0/32.2 & 65.6/32.5 & 55.2/29.6 & 67.1/43.2 & 81.3/27.5 \\
			\bottomrule
		\end{tabular}
	}
	\vspace{-15pt}
\end{table}

\noindent \textbf{Adversarial training improves end-to-end robustness.} As shown in Table \ref{tab: adv_data_var_resnet_adv_train}, the use of adversarial training makes attacks more challenging. We observe significant drops in both the FPR as well as OW-TSR, compared to the setting where a naturally trained classifier is used (Table \ref{tab: adv_data_var_resnet}). 

\begin{table}[t]
	\caption{FPR of OOD detectors for CIFAR-10 data for 6 different OOD datasets with adversarially trained models on unmodified OOD data. \textbf{Model:} WRN-28-10, \textbf{In-distribution:} CIFAR-10.
	Autoencoder and Deep-SVDD are independent of the classifier used, so have the same performance as in Table \ref{tab: benign_data_var_resnet}.}

	\label{tab: benign_data_var_resnet_adv_train}
	\resizebox{\linewidth}{!}{
		\begin{tabular}{c|c|c|c|c|c|c}
			\toprule
			\multirow{2}{*}{OOD Detector}  & \multirow{2}{*}{MNIST} & \multirow{2}{*}{VOC12} & \multirow{2}{*}{Imagenet} & Internet  & Gaussian  & Uniform  \\ 
              &  &  &  &  Photos &  Noise &  Noise \\ \midrule
			ODIN &  77.9 & 65.3 & 72.8 & 70.9 & 99.9 & 100.0 \\ \hline
			Network Agnostophobia & 55.2  & 61.4 & 58.6 & \textbf{52.9} & \textbf{0.0} & \textbf{0.0} \\ \hline
			Mahalanobis Detector & 9.4  & 63.5 & \textbf{46.0} & 73.1 & \textbf{0.0} & \textbf{0.0} \\ \hline
			Outlier Exposure & 71.9 & 68.7 & 68.6 & 57.6 & 93.7 & 99.9 \\
			\bottomrule
		\end{tabular}
	}
	\vspace{-15pt}
\end{table}

\begin{figure}[t]
	\centering
	\subfloat[Benign classifier]{\resizebox{0.28\textwidth}{!}{\includegraphics[width=\linewidth]{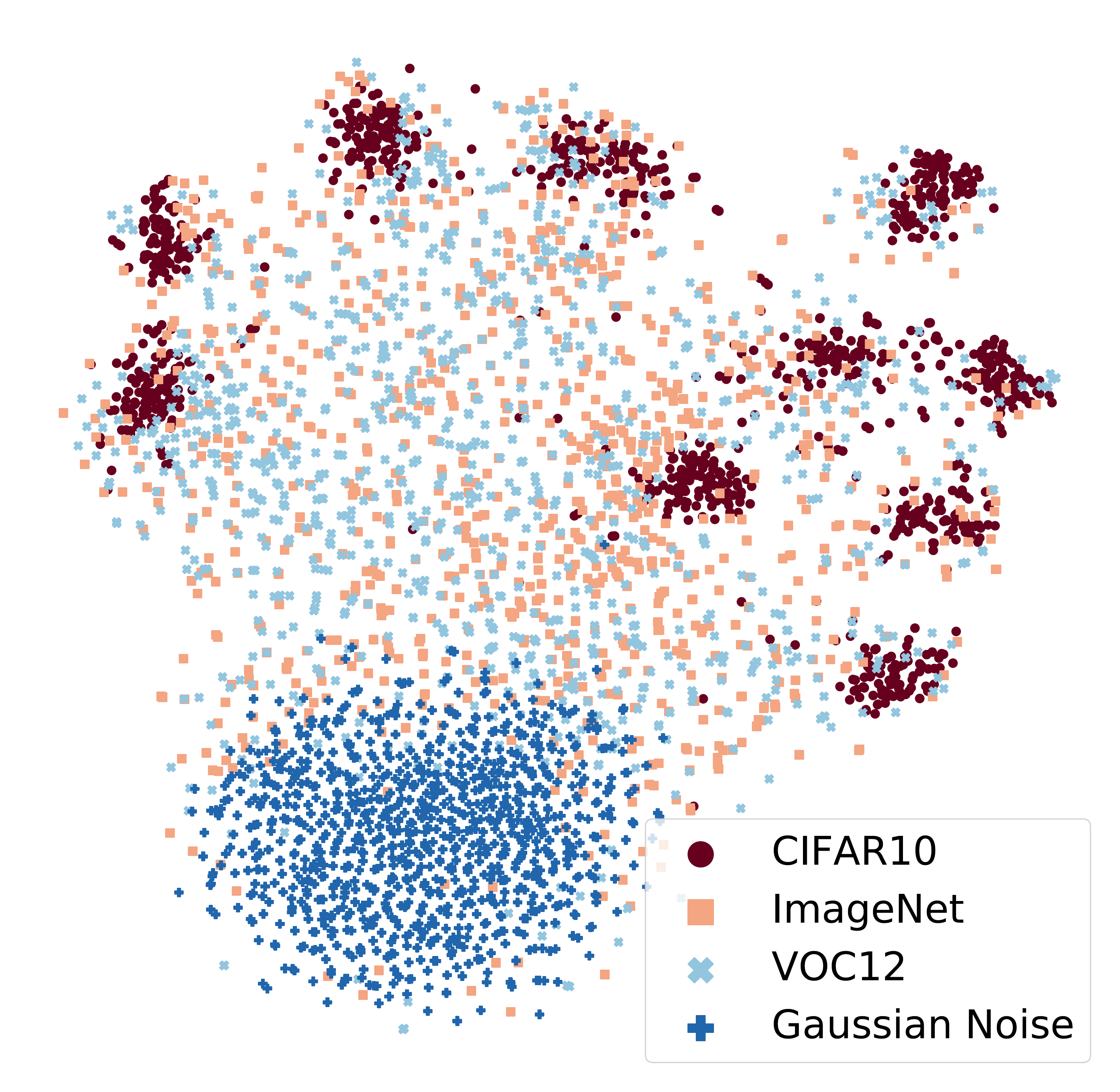}}\label{subfig: embed_tsne_ben}}
	\hspace{2mm}
	\subfloat[Robustly trained classifier]{\resizebox{0.28\textwidth}{!}{\includegraphics[width=\linewidth]{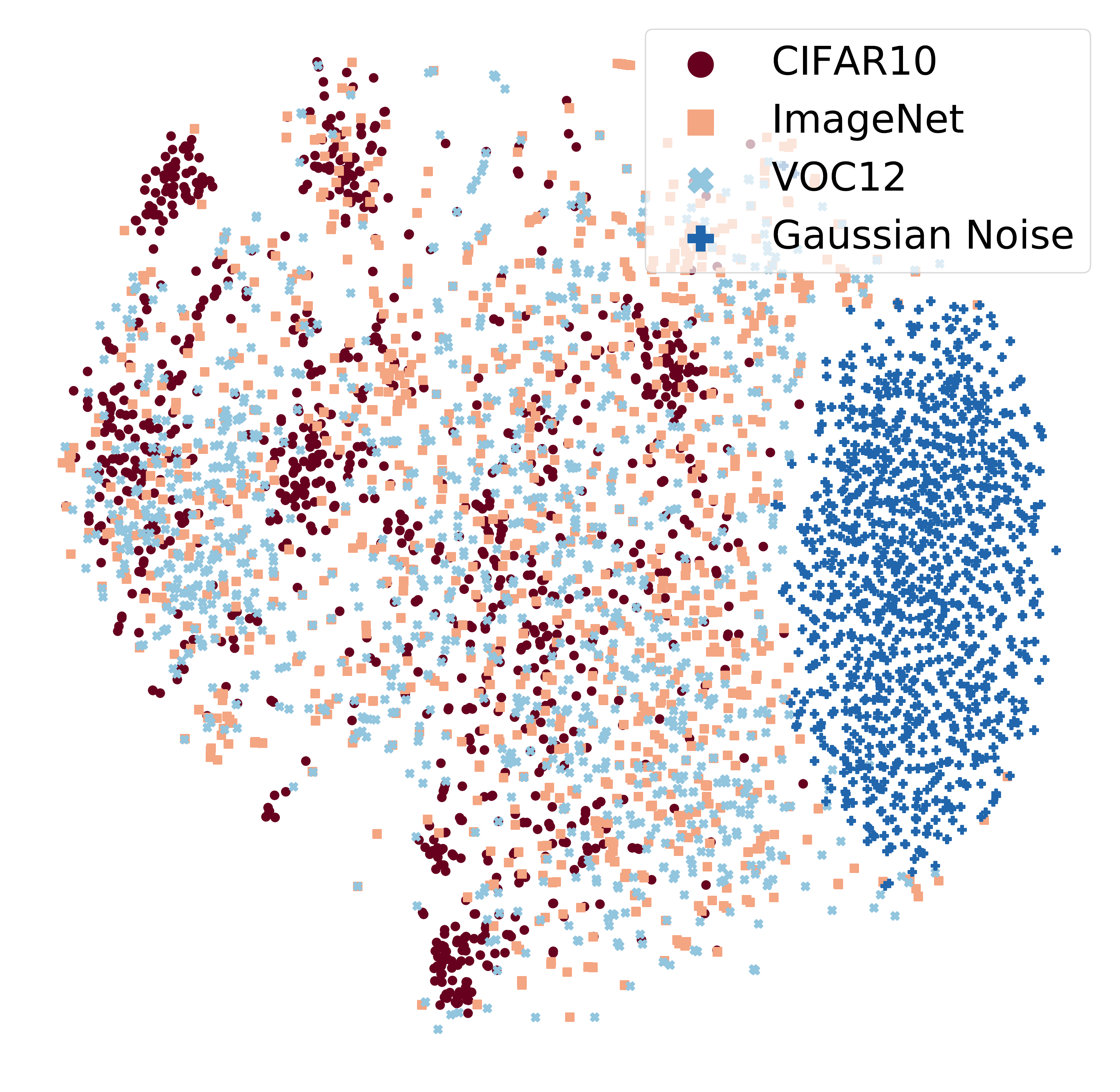}}\label{subfig: embed_tsne_adv}}
	\caption{t-SNE visualization of embedding of the last convolutional layer of WRN-28-10 trained on CIFAR-10.}
	\label{fig: tsne_viz}
	\vspace{-10pt}
\end{figure}

\noindent \textbf{Adversarial training reduces discriminative power and leads to worse detection performance on benign OOD data.} We find that adversarial training does have a severe deleterious impact on benign OOD detection (Table \ref{tab: benign_data_var_resnet_adv_train}).
In Figure \ref{fig: tsne_viz}, we investigate it further by visualizing the last convolutional layer's feature representations of both naturally and adversarially trained classifiers with the t-SNE \cite{maaten2008visualizing} method.
In Fig. \ref{subfig: embed_tsne_ben}, $D_{\text{in}}$ (CIFAR-10) is clearly separated from the other datasets. 
However, this is not the case in Fig. \ref{subfig: embed_tsne_adv}, where representations from $D_{\text{in}}$ overlap considerably with those from OOD data.

\section{Concluding Remarks} \label{sec: discussion}
In this paper, we have performed a thorough empirical analysis of open-world ML systems for unmodified, corrupted and adversarially perturbed \ood data. 
We believe our work provides a template for future evaluations of robust OOD detectors and open-world ML systems and make the following recommendations. 
First, detector performance must be checked across different network architectures and in-distribution datasets. 
Second, OOD detectors must display consistent behavior across different OOD datasets, as well as corrupted versions of these. 
Finally, open-world ML systems must be designed to be robust to adaptive attackers and must be evaluated on OOD adversarial examples.
\section*{Acknowledgements}
This work was supported in part by the National Science Foundation under grants CNS-1553437 and CNS-1704105, the Office of Naval Research Young Investigator Award, the Army Research Office Young Investigator Prize, Faculty research award from Facebook, and by Schmidt DataX award.

\bibliography{references}
\bibliographystyle{icml2020}

\appendix
\section{Tiny-ImageNet results}\label{appendix:tiny-imagenet}
\vspace{-5pt}
\begin{table}[ht]
	\caption{FPR (lower is better) of OOD detectors for 6 different OOD datasets at a 95\% TPR. \textbf{Model:} WRN-28-10, \textbf{In-distribution:} Tiny-ImageNet.}
	    \resizebox{\linewidth}{!}{
		\begin{tabular}{c|c|c|c|c|c}
		    \toprule
            \multirow{2}{*}{OOD Detector}  & \multirow{2}{*}{MNIST} & \multirow{2}{*}{VOC12}  & Internet & Gaussian & Uniform \\
             & &  & Photos &  Noise & Noise\\ \midrule
            ODIN & \textbf{0.0}  & \textbf{86.4} & 88.7 & 100 & 100 \\ \hline
            Mahalanobis Detector & \textbf{0.0}  & 92.3  & \textbf{67.3} & \textbf{0.0} & \textbf{0.0} \\ 
            \bottomrule
        \end{tabular}
	}
	\label{tab:imagenet-odin}
	\vspace{-20pt}
\end{table}
\begin{table}[ht]
	\caption{FPR of OOD detectors for 6 different OOD datasets with adversarially trained models on unmodified OOD data. \textbf{Model:} WRN-28-10, \textbf{In-distribution:} Tiny-ImageNet.}
	\resizebox{\linewidth}{!}{
		\begin{tabular}{c|c|c|c|c|c}
			\toprule
			\multirow{2}{*}{OOD Detector}  & \multirow{2}{*}{MNIST} & \multirow{2}{*}{VOC12}  & Internet & Gaussian & Uniform \\
             & &  & Photos &  Noise & Noise\\ \midrule
			ODIN & 100  & \textbf{88.9} & 90.6 & 100 & 100  \\ \hline
			Mahalanobis Detector& \textbf{0.0}  & 98.1  & \textbf{82.4} & \textbf{0.0} & \textbf{0.0} \\
			\bottomrule
		\end{tabular}
	}
	\vspace{-7pt}
	\label{tab:imagenet-mahalanobis}
\end{table}

Table \ref{tab:imagenet-odin} and Table \ref{tab:imagenet-mahalanobis} present the detection performance of ODIN and Mahalanobis Detector on both naturally and adversarially trained Tiny-ImageNet classifiers.
We find that both detectors return very high FPR when using VOC12 and Internet Photos as OOD datasets.
We also notice that using adversarially trained classifier leads to larger FPR values.
\section{All corruption results}\label{appendix:corruptions}
\begin{figure*}[ht]
	\centering
	\subfloat[ODIN detector (geometric)]
	{\resizebox{0.4\textwidth}{!}{\includegraphics[width=0.6\linewidth]{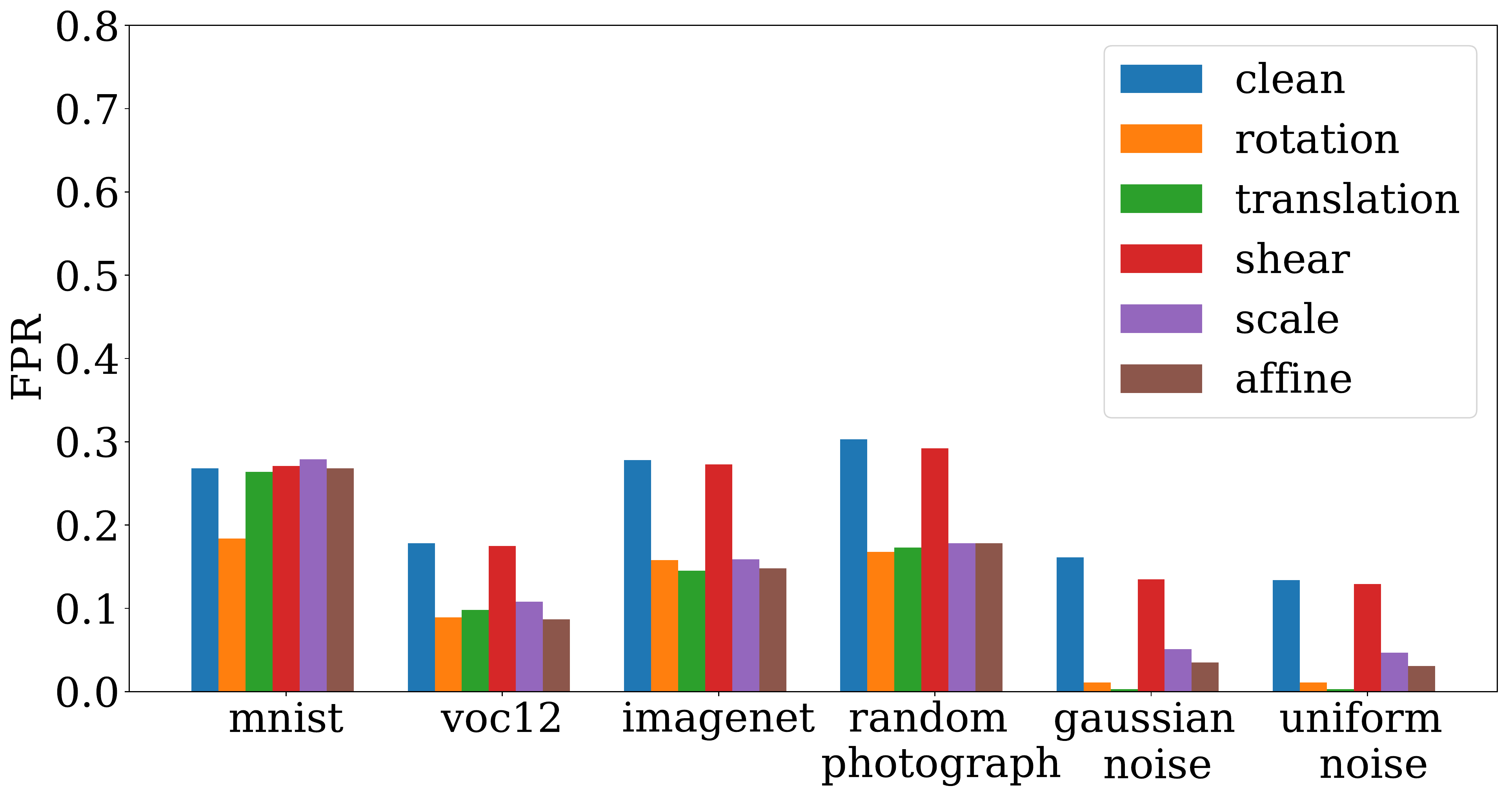}}}
	\hspace{0mm}
	\subfloat[Mahalanobis detector (geometric)]
	{\resizebox{0.4\textwidth}{!}{\includegraphics[width=0.6\linewidth]{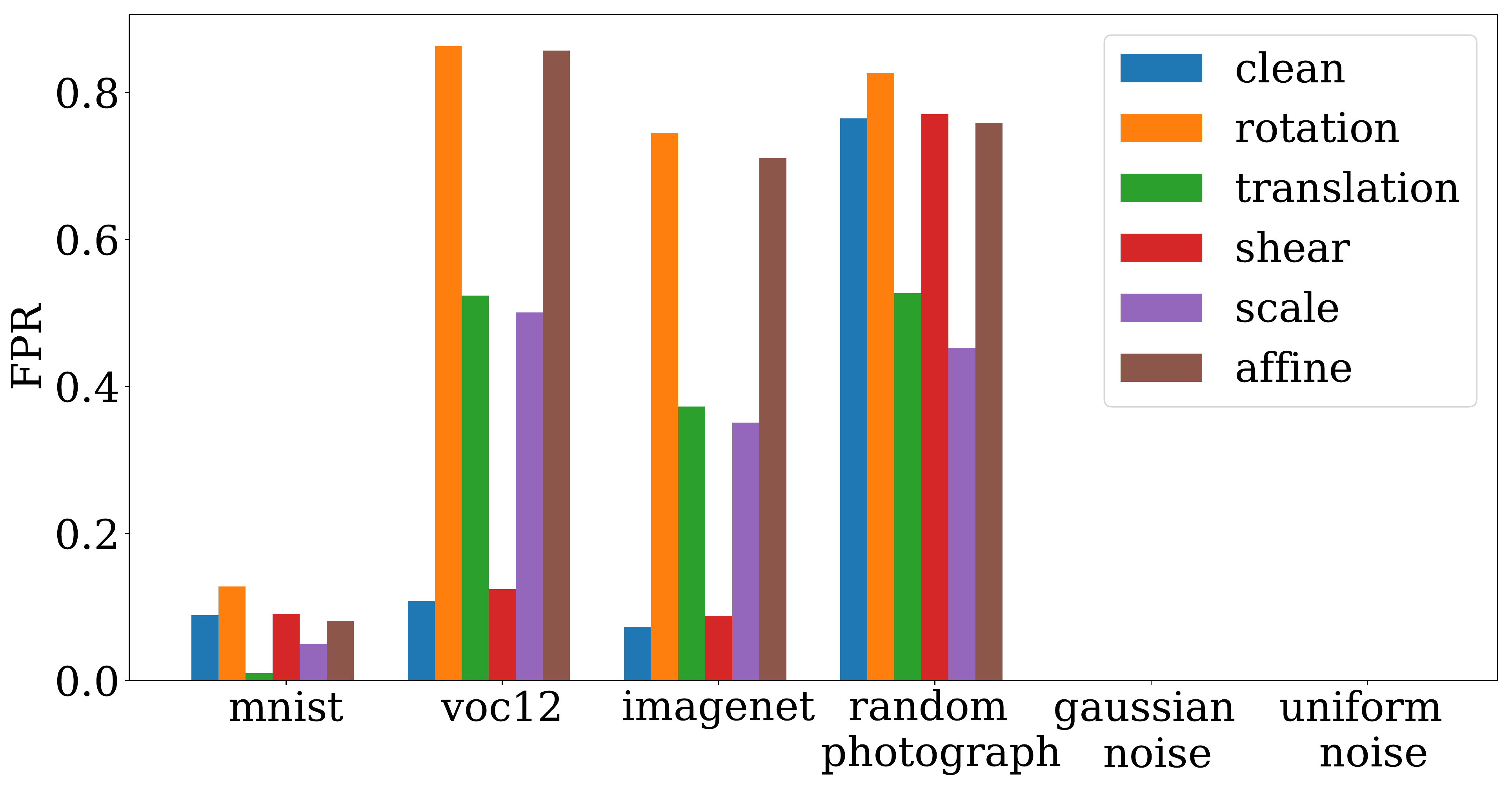}}}
	\hspace{0mm}
	\subfloat[ODIN detector (blurring)]
	{\resizebox{0.4\textwidth}{!}{\includegraphics[width=0.6\linewidth]{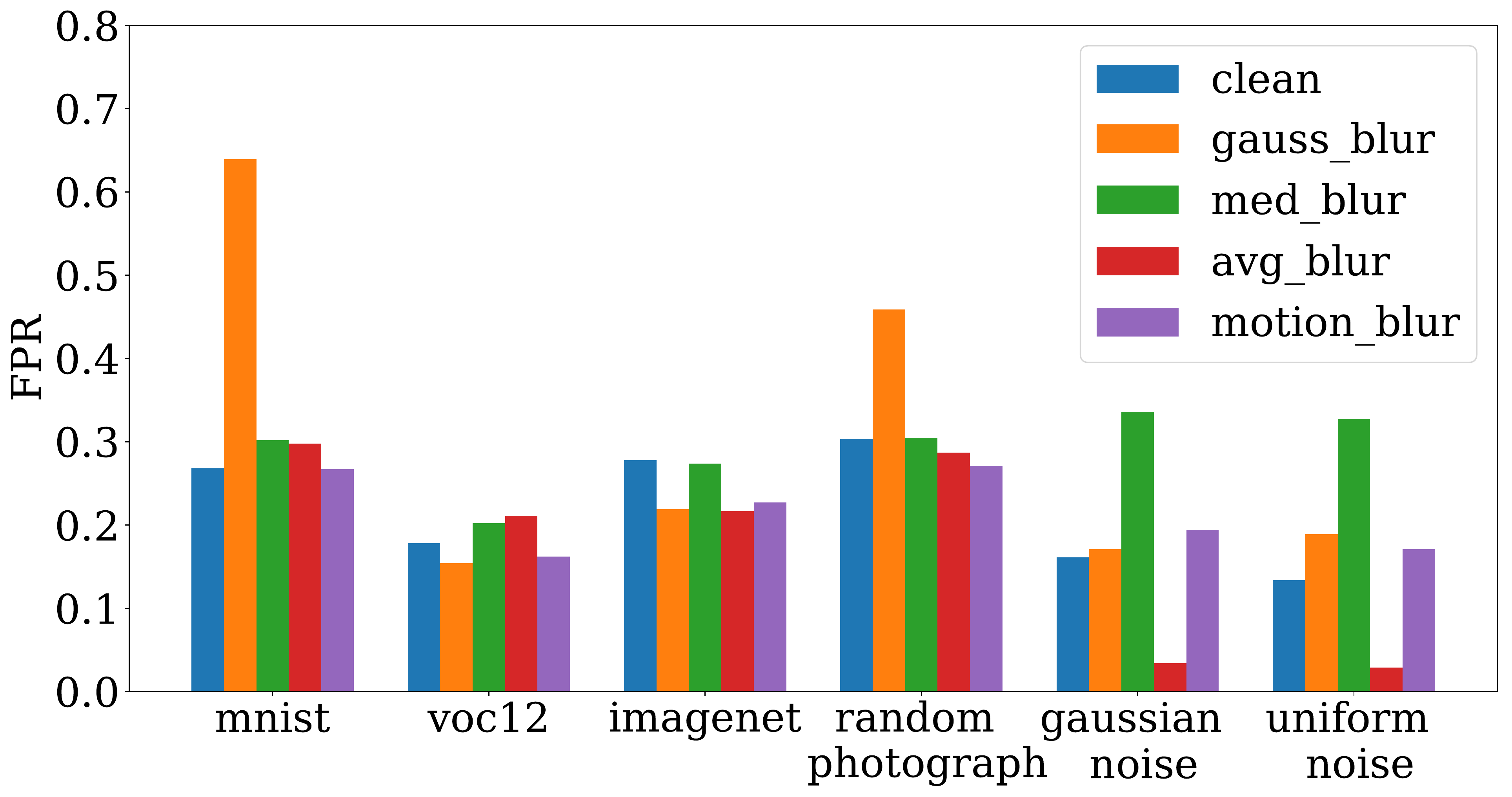}}}
	\hspace{0mm}
	\subfloat[Mahalanobis detector (blurring)]
	{\resizebox{0.4\textwidth}{!}{\includegraphics[width=0.6\linewidth]{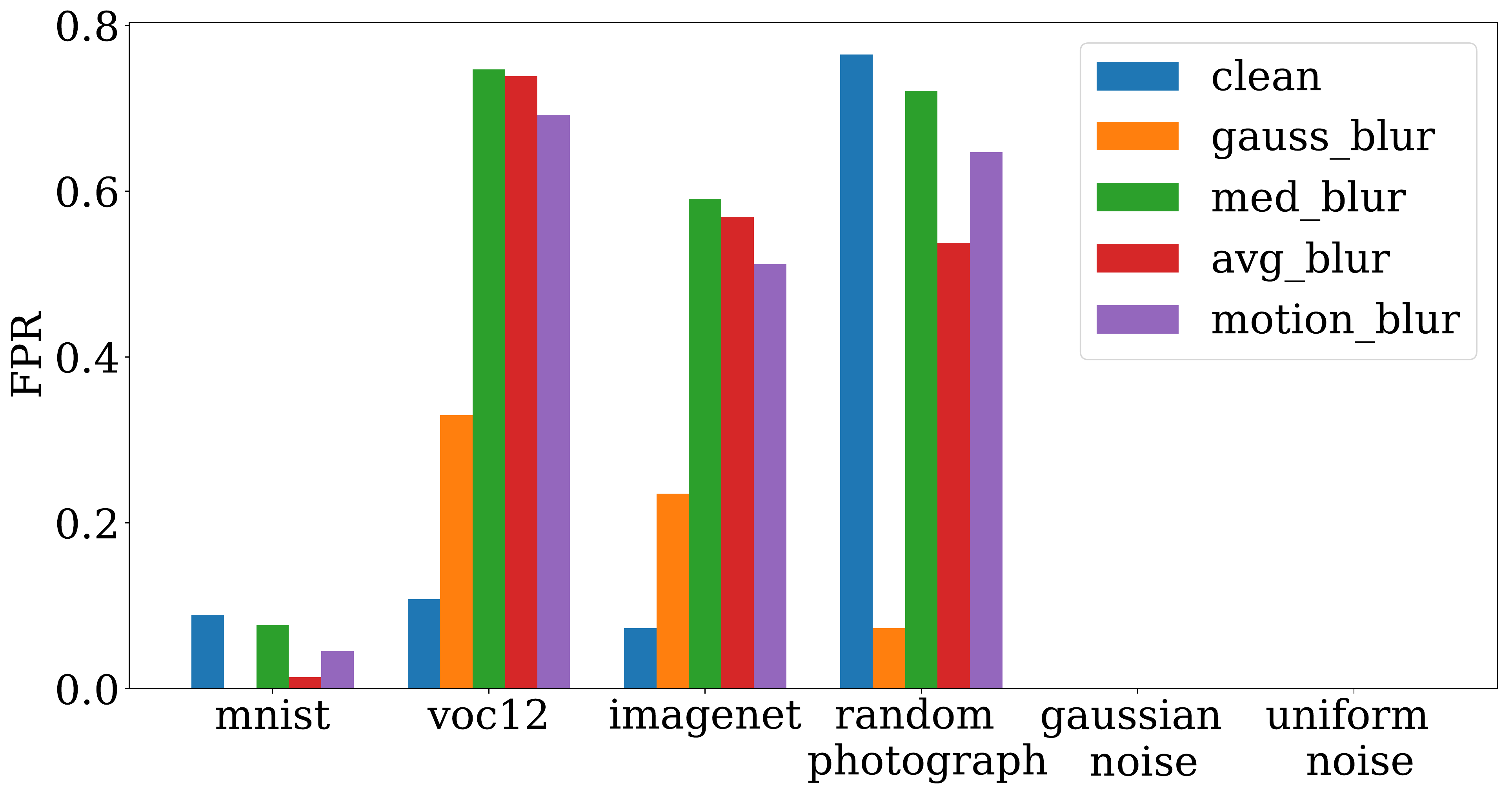}}}
	\hspace{0mm}
	\subfloat[ODIN detector (noise)]
	{\resizebox{0.4\textwidth}{!}{\includegraphics[width=0.6\linewidth]{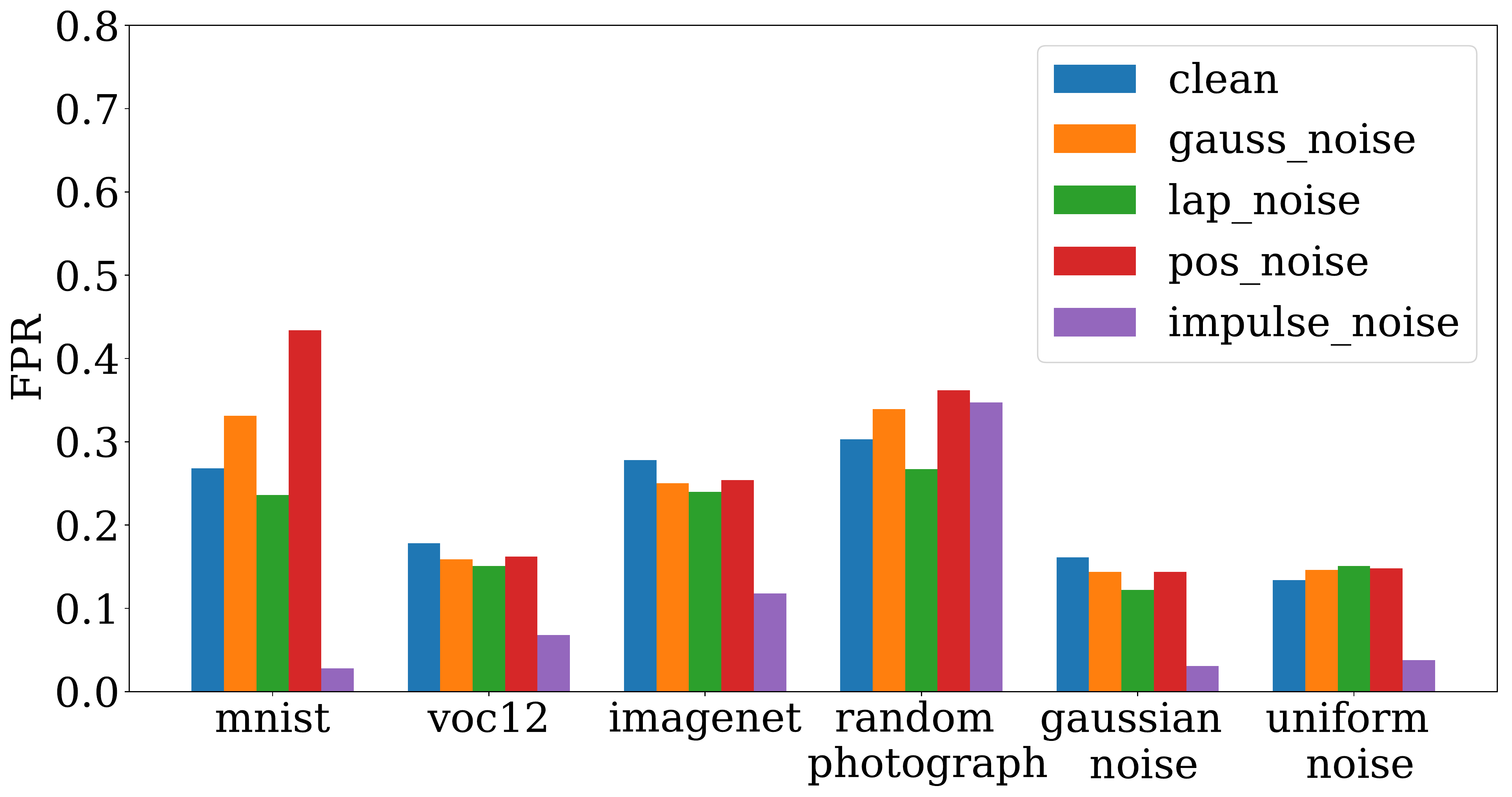}}}
	\hspace{0mm}
	\subfloat[Mahalanobis detector (noise)]
	{\resizebox{0.4\textwidth}{!}{\includegraphics[width=0.6\linewidth]{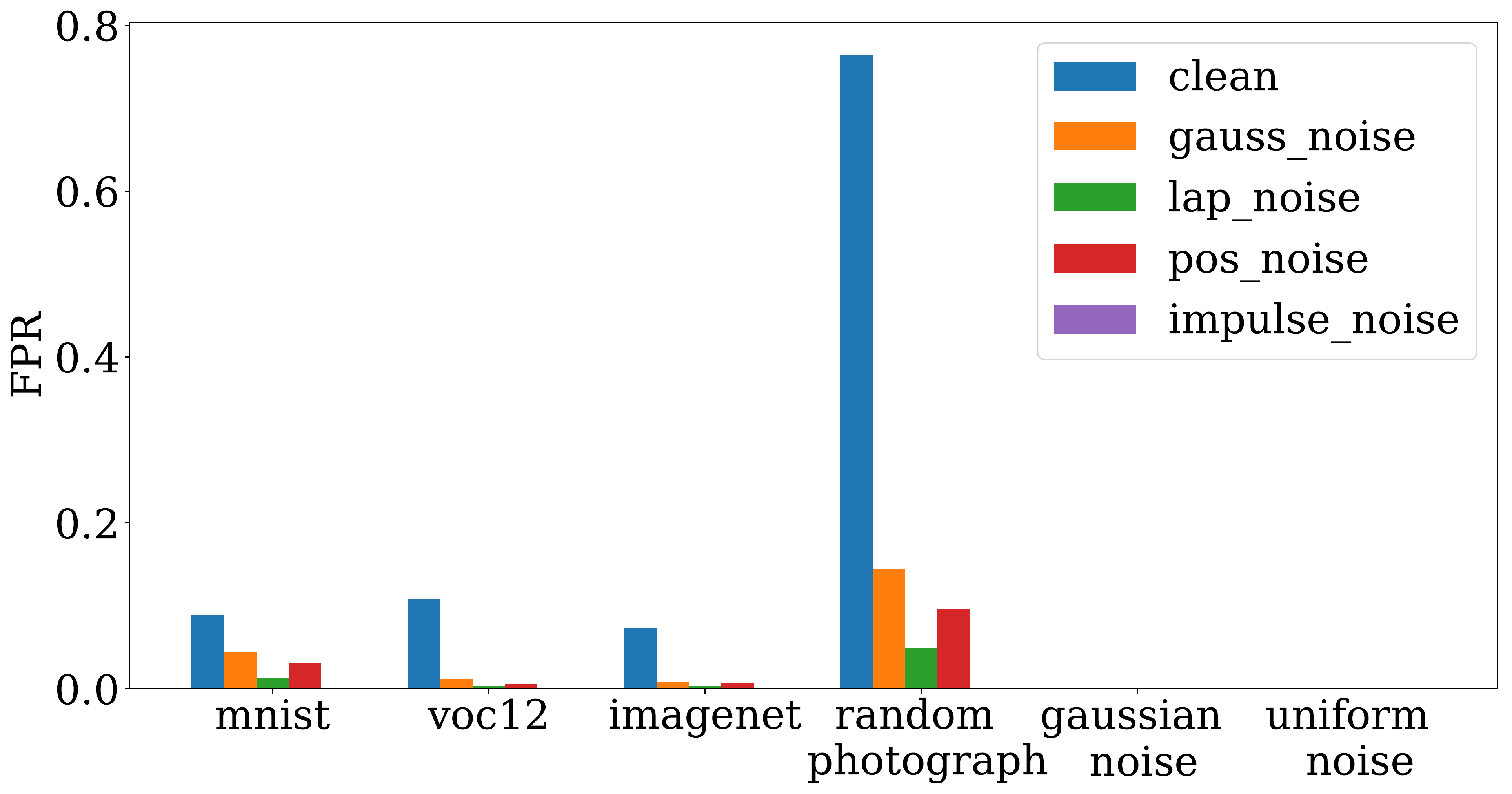}}}
	\hspace{0mm}
	\subfloat[ODIN detector (photometric)]
	{\resizebox{0.4\textwidth}{!}{\includegraphics[width=0.6\linewidth]{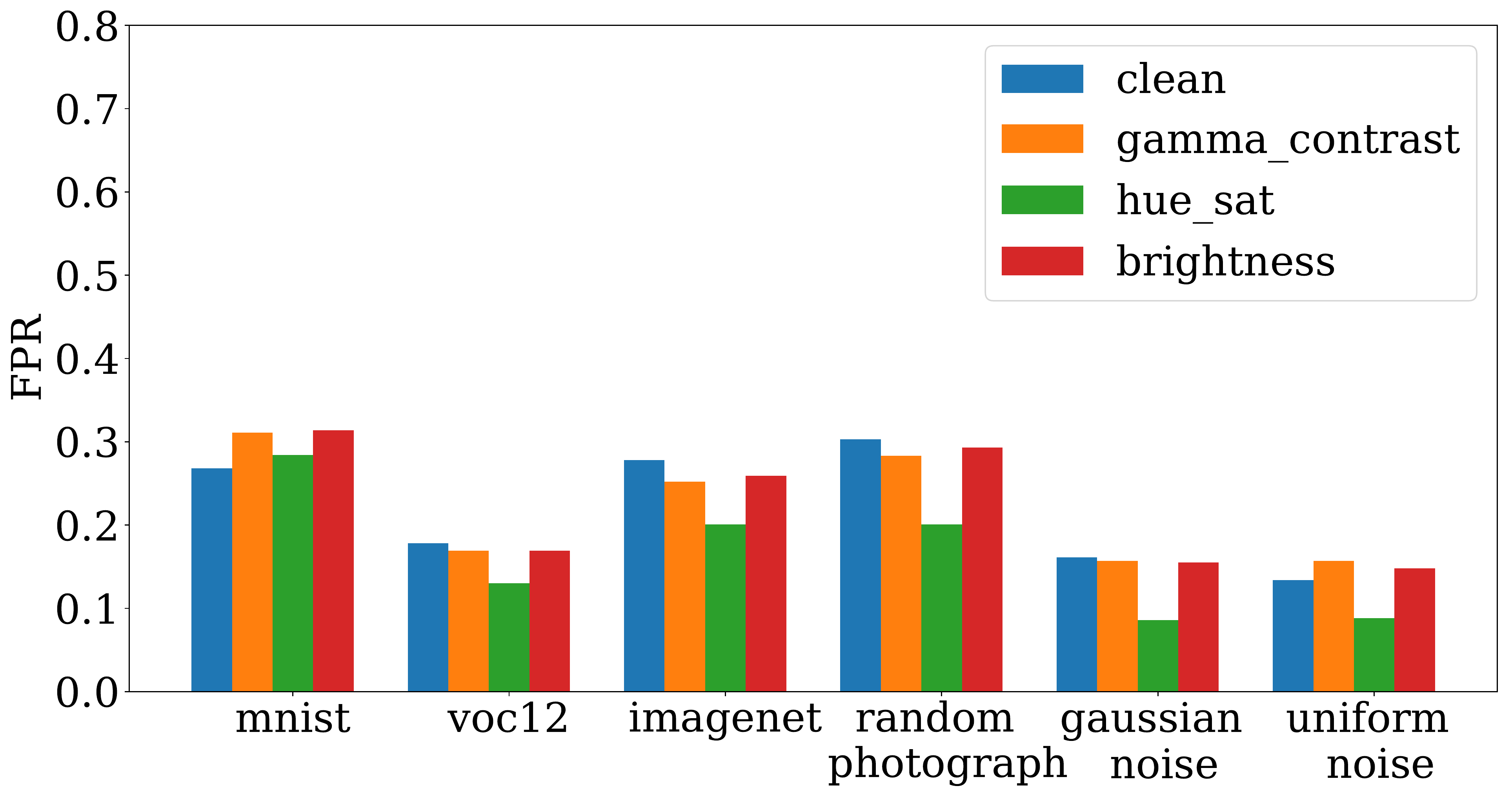}}}
	\subfloat[Mahalanobis detector (photometric)]
	{\resizebox{0.4\textwidth}{!}{\includegraphics[width=0.6\linewidth]{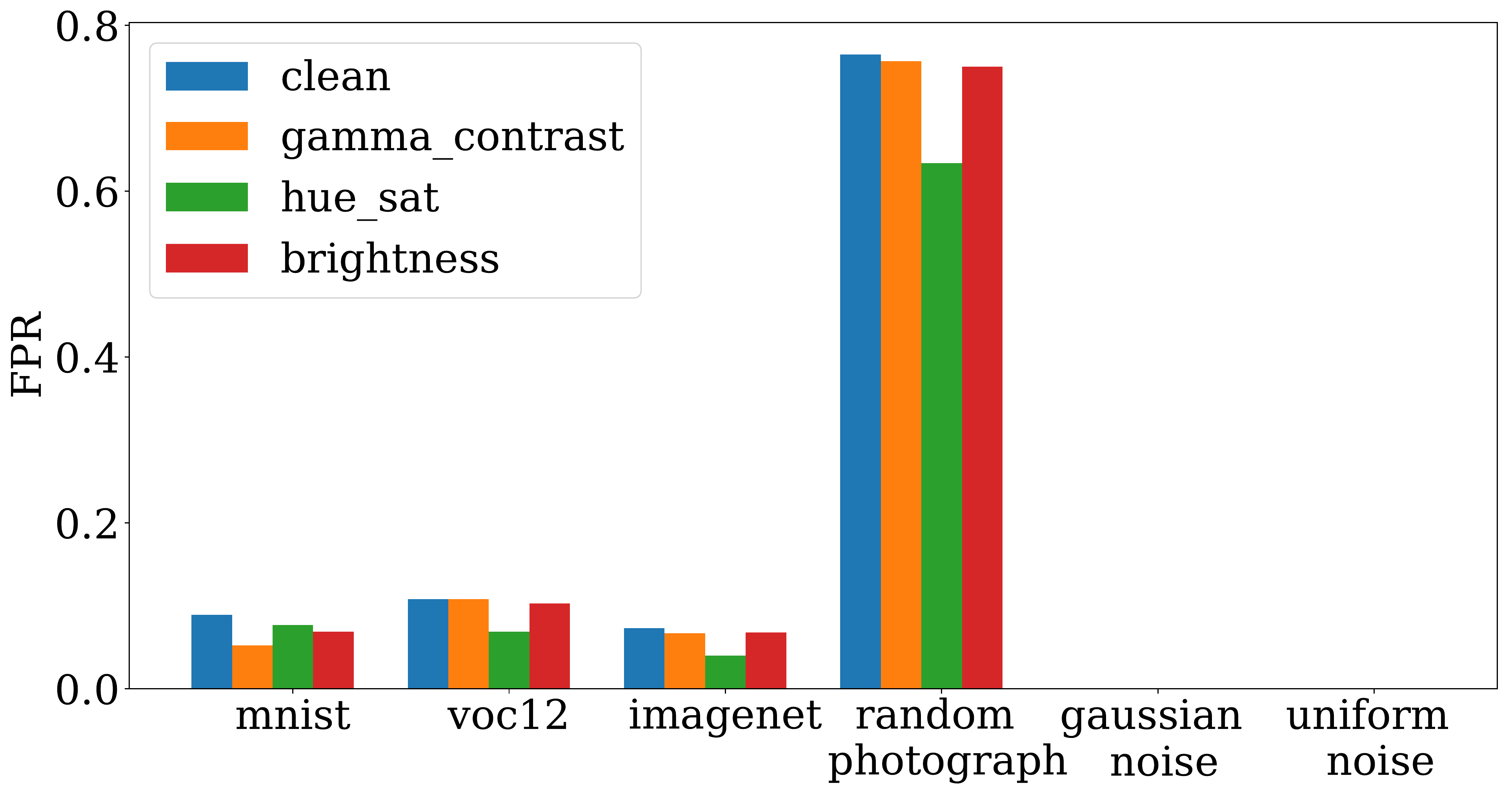}}}
	\hspace{0mm}
	\subfloat[ODIN detector (weather)]
	{\resizebox{0.4\textwidth}{!}{\includegraphics[width=0.6\linewidth]{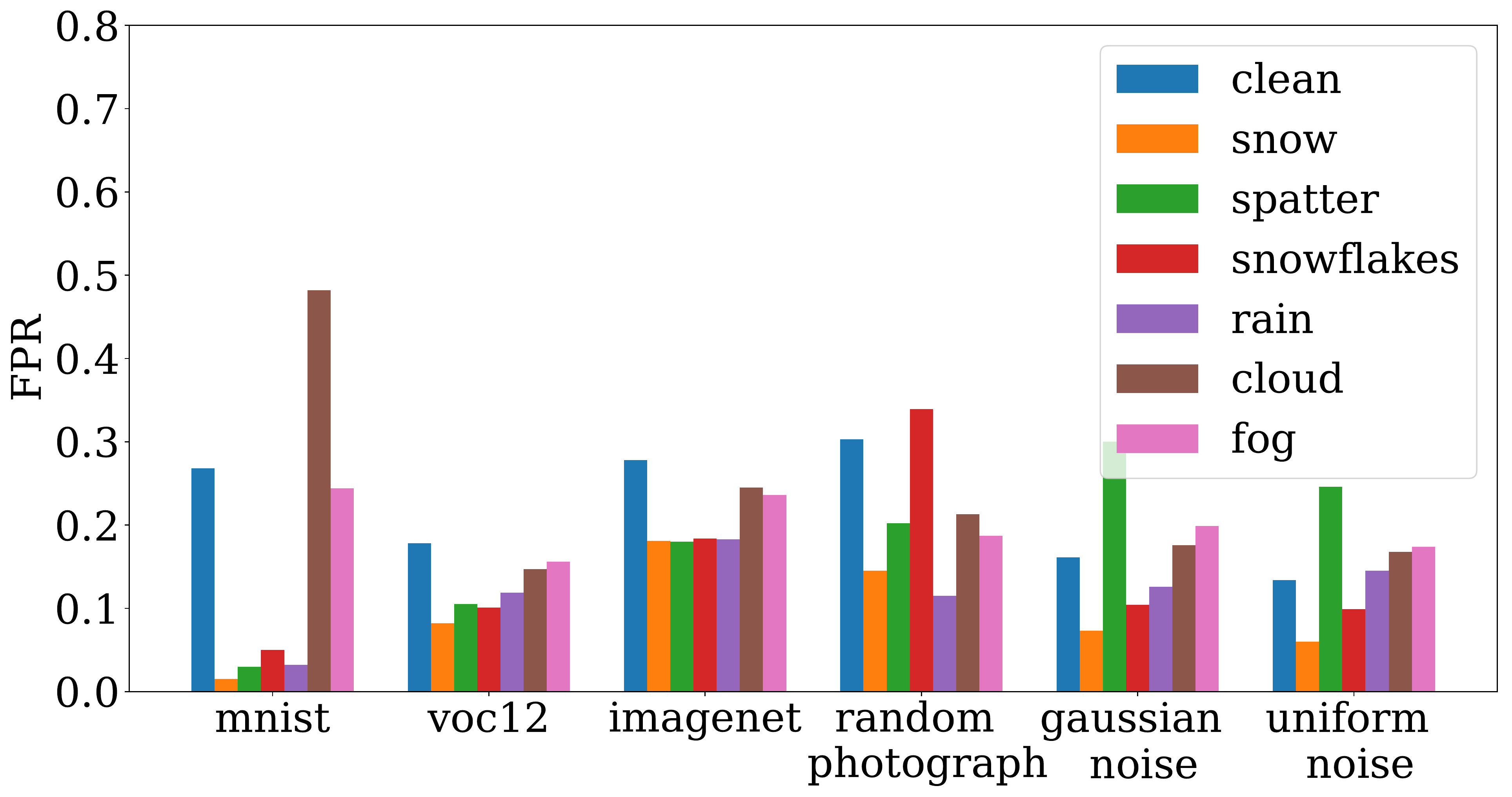}}}
	\subfloat[Mahalanobis detector (weather)]
	{\resizebox{0.4\textwidth}{!}{\includegraphics[width=0.6\linewidth]{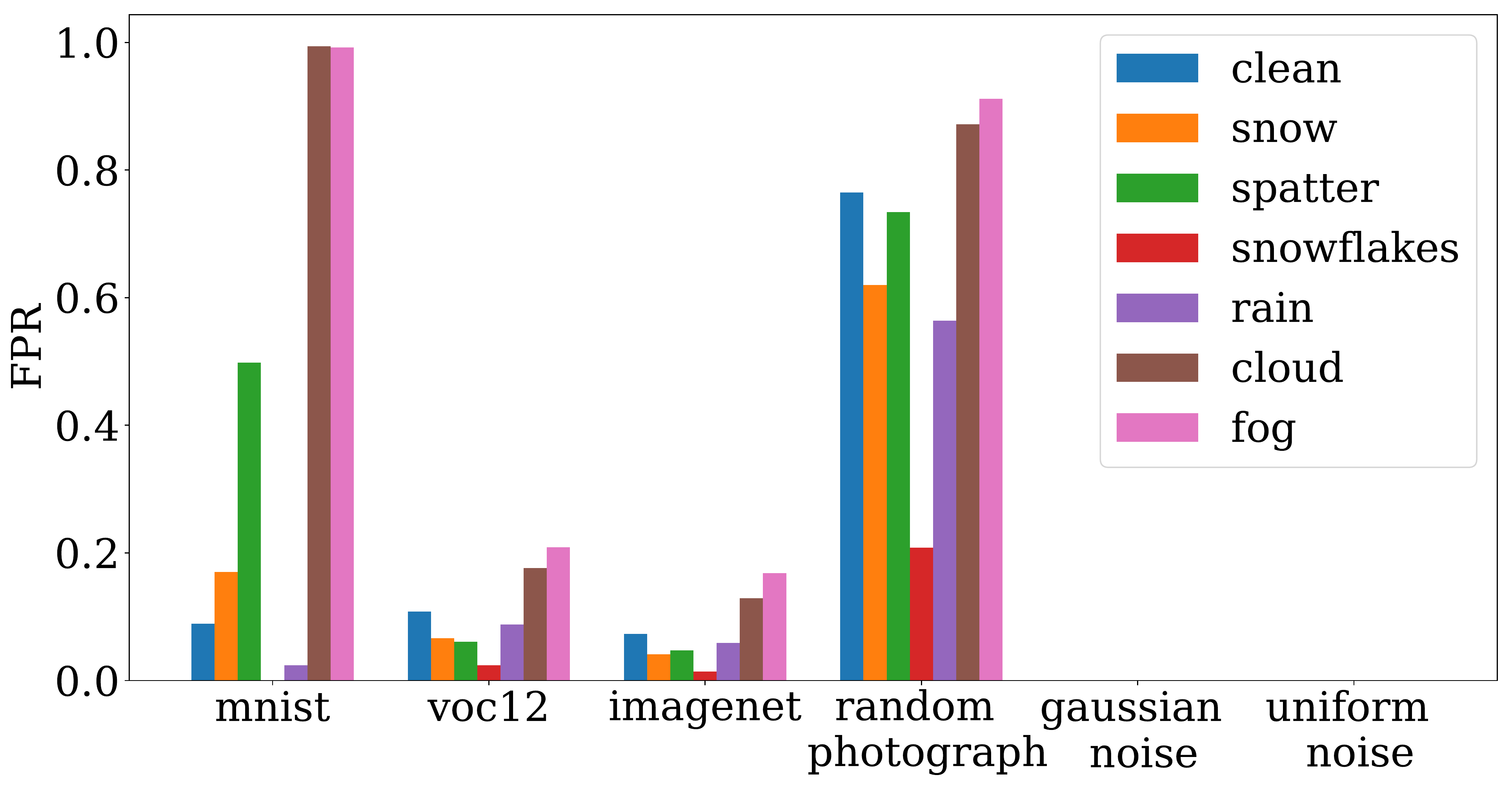}}}
	\caption{FPR for OOD detectors on corrupted OOD data, the first column is for ODIN, the second column is for the Mahalanobis detector. Each row represents geometric, blurring, noise, photometric, and weather corruptions, respectively. \textbf{Model:} WRN-28-10, \textbf{In-distribution:} CIFAR-10.}
	\label{fig: cifar_corruption_appendix}
\end{figure*}

Figure \ref{fig: cifar_corruption_appendix} shows the detection performance of the ODIN and Mahalanobis detectors on the natural CIFAR10 classifiers with all 22 image corruptions.
We find that certain corruptions leads to higher FPR values: all geometric and blurring corruptions increase FPR values for VOC12 or ImageNet. On the other hand, noise-based corruptions have a tendency to \emph{decrease} FPR rates.

\end{document}